\documentclass[letterpaper, 10 pt, conference]{ieeeconf}  % Comment this line out if you need a4paper

\IEEEoverridecommandlockouts                              % This command is only needed if 
\usepackage{amsmath} % assumes amsmath package installed
\usepackage{amssymb}  % assumes amsmath package installed
\usepackage{hyperref}

\title{\LARGE \bf
\fullname: In-Hand, Contact-Rich, and Long-Horizon Dexterous Robot Drumming
}

\author{Hung-Chieh Fang, Amber Xie, Jennifer Grannen,  Kenneth Llontop, Dorsa Sadigh% <-this % stops a space
\\
Stanford University
\thanks{Work was partially done during Hung-Chieh Fang’s internship at Stanford University. Hung-Chieh is now with National Taiwan University.}%
}

\usepackage{xcolor} % for colorbox
\usepackage{graphicx} % for resizebox
\usepackage{booktabs}
\usepackage{cleveref} 
\usepackage{multirow}
\usepackage{dsfont}
\usepackage{colortbl}
\usepackage{xspace}

\newcommand{\fullname}[0]{DexDrummer\xspace}

\begin{document}

\bstctlcite{IEEEexample:BSTcontrol}
\maketitle
\thispagestyle{empty}
\pagestyle{empty}

%%%%%%%%%%%%%%%%%%%%%%%%%%%%%%%%%%%%%%%%%%%%%%%%%%%%%%%%%%%%%%%%%%%%%%%%%%%%%%%%
\begin{abstract}
Performing in-hand, contact-rich, and long-horizon dexterous manipulation remains an unsolved challenge in robotics. Prior hand dexterity works have considered each of these three challenges in isolation, yet do not combine these skills into a single, complex task. 
To further test the capabilities of dexterity, we propose drumming as a testbed for dexterous manipulation. Drumming naturally integrates all three challenges: it involves in-hand control for stabilizing and adjusting the drumstick with the fingers, contact-rich interaction through repeated striking of the drum surface, and long-horizon coordination when switching between drums and sustaining rhythmic play. 
We present \fullname, a hierarchical object-centric bimanual drumming policy trained in simulation with sim-to-real transfer. The framework reduces the exploration difficulty of pure reinforcement learning by combining trajectory planning with residual RL corrections for fast transitions between drums. A dexterous manipulation policy handles contact-rich dynamics, guided by rewards that explicitly model both finger–stick and stick–drum interactions.
In simulation, we show our policy can play two styles of music: multi-drum, bimanual songs and challenging, technical exercises that require increased dexterity. Across simulated bimanual tasks, our dexterous, reactive policy outperforms a fixed grasp policy by 1.87x across easy songs and 1.22x across hard songs F1 scores. In real-world tasks, we show song performance across a multi-drum setup.  \fullname is able to play our training song and its extended version with an F1 score of 1.0. Code and videos are available at \url{https://dexdrummer.github.io/}.

\end{abstract}

\section{INTRODUCTION}
Dexterous hand manipulation is an attractive problem in robotics because it unlocks a broad set of real-world tasks. Existing works have tackled challenges such as in-hand object reorientation~\cite{openai2019solvingrubikscuberobot, wang2024penspin}, grasping~\cite{li2022gendexgrasp, zhong2025dexgrasp, miller2004graspit, wang2022dexgraspnet, brahmbhatt2019contactgrasp, xu2023unidexgrasp, zhong2025dexgraspvlavisionlanguageactionframeworkgeneral}, and tool-based manipulation~\cite{shaw2024bimanual}, all of which require managing complex finger–object interactions.
While these works provide useful insights on dexterity, these studies typically emphasize short-horizon tasks or narrow aspects of dexterity in isolation.

In contrast, many real-world tasks such as assembly or cooking require dexterous skills that combine in-hand control, robustness to external perturbations, and long-horizon robustness. For example, assembling parts often involves reorienting a fastener in the hand while applying force to connect components, and cooking requires both holding utensils stably and stirring against resistance.

Motivated by the need for a compelling testbed, we propose drumming, a long-horizon, contact-rich dexterous manipulation task. Drumming inherently requires balancing in-hand control -- maintaining and adjusting the grasp of the stick with fine finger control -- and external contact -- forcefully and repeatedly striking drums. To play long songs, this control becomes even more crucial: drumming requires a policy robust to these contacts for extended periods of time. 

\begin{figure*}
\vspace{4mm}
    \centering

\includegraphics[width=0.9\linewidth]{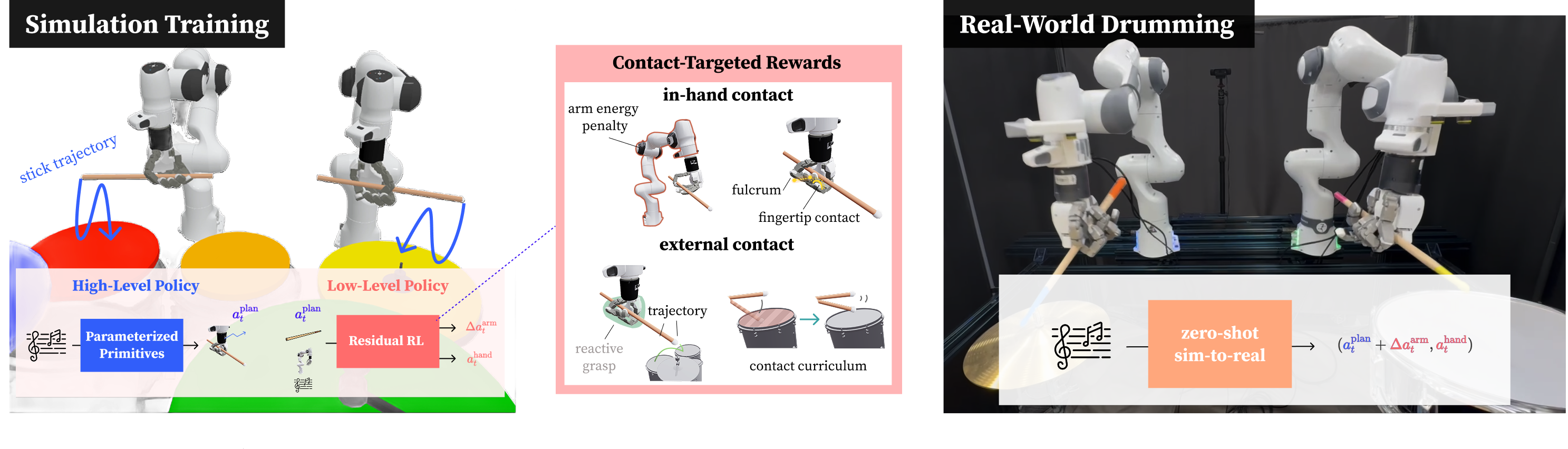}
\vspace{-10pt}
    
    \caption{\textbf{\fullname}: A hierarchical dexterous drumming framework trained in simulation and deployed in the real world.
\textbf{Left: Training in simulation.} Our policy decomposes the task into a hierarchical controller. A \emph{high-level policy} generates parameterized motion primitives that produce drumstick trajectories from musical inputs. A \emph{low-level policy} then uses residual RL to learn corrective arm and hand actions for accurate trajectory tracking during fast transitions.
\textbf{Middle: Contact-targeted rewards.} To handle contact-rich dynamics, we design rewards that target two types of interactions. \emph{In-hand contacts} encourage stable finger–stick manipulation through fingertip contact rewards, a fulcrum reward between the thumb and index finger, and an arm energy penalty that promotes finger-dominant control. \emph{External contacts} address stick–drum interactions with a trajectory guidance reward to encourage drum strikes and a contact curriculum that gradually introduces drum contact during training.
\textbf{Right: Real-world drumming.} Policies trained in simulation transfer zero-shot to the real robot, enabling dexterous multi-drum playing over long horizons.
}
    \label{fig:main}
\end{figure*}

To address the challenges of in-hand control, external forces, and long-horizon robustness, we propose \fullname, a hierarchical dexterous drumming framework trained in simulation with sim-to-real transfer. As illustrated in Fig.~\ref{fig:main}, our system decomposes the task into a high-level policy and a low-level dexterous policy.

At the high level, we reduce the exploration difficulty of pure reinforcement learning by introducing parameterized motion primitives that generate task-space drumstick trajectories from musical inputs. These trajectories are converted into arm motions via motion planning, producing nominal control commands for the robot arms. A residual RL policy then learns corrective adjustments on top of this planner to compensate for tracking errors during fast transitions between drums.

At the low level, we train a dexterous manipulation policy that handles the contact-rich dynamics of drumming. Our key insight is to structure learning using contact-targeted rewards, which explicitly address two types of interactions: in-hand contacts and external contacts. In-hand contacts correspond to finger–stick interactions that manipulate the drumstick through fingertip contact and a fulcrum grasp and stabilize through arm energy penalty. External contacts correspond to interactions between the stick and the drum surface. To learn robust striking behavior, we introduce trajectory-guided rewards and a contact curriculum that stabilizes learning of impacts.

In simulation, we show \fullname{} performance on two styles of music: bimanual, multi-drum songs, and a uni-drum technical exercise. We show sim-to-real transfer for real-world drumming, showcasing \fullname's capability to perform dexterous, contact-rich control of the drum stick across a long horizon.

In summary, our main contributions are threefold:

\begin{enumerate}
    \item We introduce drumming as a challenging testbed for dexterous manipulation, unifying the challenges of in-hand manipulation, external contacts, and long-horizon robustness.
    \item We propose \fullname, a hierarchical, two-stage policy for drumming. \fullname consists of a high-level residual RL policy and a low-level dexterous policy that optimized contact-targeted rewards, to address the challenges of in-hand and external contacts. 
    \item We showcase simulated drumming across six different musical genres at two difficulty levels, with performances ranging from 20 to 40 seconds, as well as a high-speed exercise piece designed to test finger dexterity. In addition, we evaluate our method on a real-world drumming task using a cymbal and snare setup.
\end{enumerate}

\section{Related Work}

\textbf{Dexterous Manipulation.} 
Dexterous manipulation has been a long-standing challenge in robotics. 
Many prior works have focused on tasks falling under these broad categories: in-hand manipulation, grasping, and post-grasp tool use.
First, in-hand manipulation typically consists of rotating or translating objects with multi-finger control \cite{Bhatt__2021, 6907059, yin2025learninginhandtranslationusing, yin2023rotatingseeinginhanddexterity, qi2023generalinhandobjectrotation}. Next, dexterous grasping is a useful skill for automation or household tasks~\cite{zhong2025dexgrasp,li2022gendexgrasp,lum2024gripmultifingergraspevaluation}. Then, grasping an object, the robot may place the object (pick-and-place) or use it as a tool, such as for drilling~\cite{fang2025dexopdevicerobotictransfer}, pouring~\cite{shaw2024bimanual}, wiping~\cite{debakker2025scaffoldingdexterousmanipulationvisionlanguage}, scissor cutting~\cite{wang2024dexcapscalableportablemocap}, and more. While these tasks are challenging and important for researching dexterity, we propose drumming, a task that requires both in-hand control, post-grasp tool-based contact, and long-horizon robustness for playing long songs.

To learn dexterous policies, many classical methods use planning with precise models~\cite{han1998dextrous,bai2014dexterous,mordatch2012contact, xie2026handelbotrealworldpianoplaying}, and propose grasps based on collision detection or optimization-based methods~\cite{miller2004graspit,liu2020deep,brahmbhatt2019contactgrasp,liu2021synthesizing,wang2022dexgraspnet,dafle2014extrinsic}. Other works use sim-to-real training with reinforcement learning~\cite{qi2022inhandobjectrotationrapid,wang2024penspin,openai2019solvingrubikscuberobot,yang2024anyrotategravityinvariantinhandobject} for in-hand tasks, or use synthetically generated grasp datasets in simulation with learning-based methods can leverage~\cite{corona2020ganhand,rosales2011synthesizing,xu2023unidexgrasp,zhong2025dexgrasp,li2022gendexgrasp,lum2024gripmultifingergraspevaluation}. Imitation learning from human demonstrations has been used for many dexterous applications~\cite{arunachalam2022holodexteachingdexterityimmersive,qin2023anyteleop,shaw2024bimanual,lin2024learningvisuotactileskillsmultifingered,zhong2025dexgraspvlavisionlanguageactionframeworkgeneral,fang2025dexopdevicerobotictransfer}. However, to reduce the data burden on designing simulated environments or teleoperating dexterous data, many works retarget human data~\cite{tao2025dexwild,xu2025dexumiusinghumanhand,wang2024dexcapscalableportablemocap}, which may also utilize reinforcement learning in simulation~\cite{agarwal2023dexterous,shaw_videodex,lum2025crossinghumanrobotembodimentgap,lin2025sim,chenobject}. 
% \hc{gap from prior work}
Following many prior works, we train an RL policy in sim, which reduces the need to hire expert drummers to generate demonstration data. We use simple sim-to-real techniques, such as domain randomization, to showcase real-world drumming.

\textbf{Robot Drumming.} Robot drumming is an exciting subset of research in robotic music-playing. Many prior works design custom hardware for drum-playing ~\cite{solenoid,afasia,6943109,berdahl2007physically}. Our project assumes a more general embodiment -- robot arm and hand -- which requires learned dexterity to hit the drums, instead of special hardware designs. Recent works explore learning-based approaches to drumming, but likewise use either custom embodiments~\cite{zrob}, or directly fix the stick to the embodiment in simulation~\cite{shahid2025drummer}, without examples of real-world drumming. Like ~\cite{shahid2025drummer}, we learn our policy in simulation via reinforcement learning, but instead of focusing on humanoid control, we assume a realistic stick-holding embodiment, which allows sim-to-real transfer. 

\begin{figure}
    \centering
    \vspace{2mm}
    \includegraphics[width=1.0\linewidth]{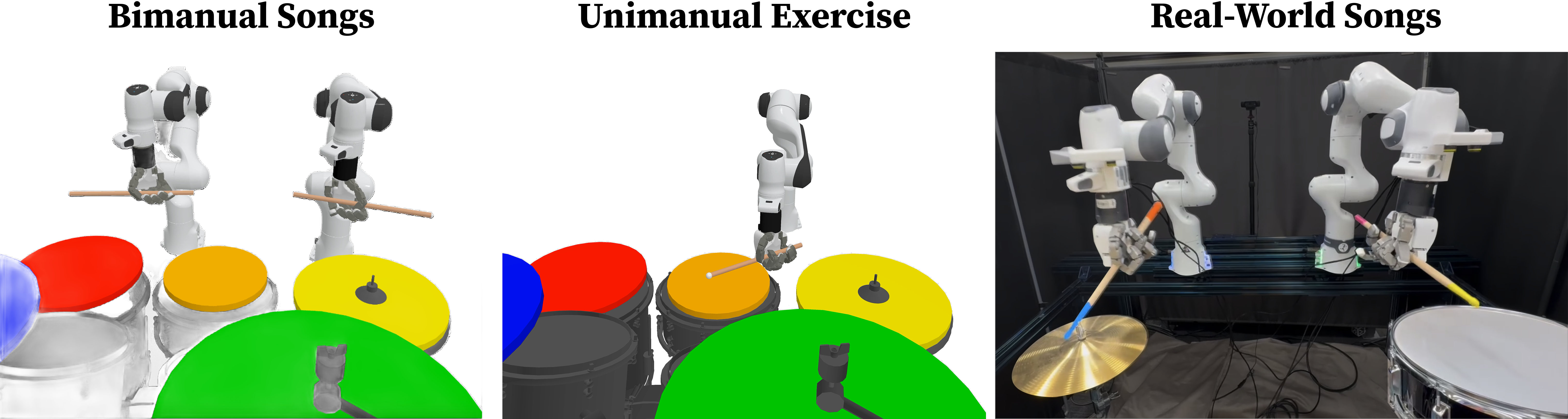}
    \caption{\textbf{Drumming Environments.} Our first simulation environment includes bimanual, multi-drum song-playing. Our second environment involves unimanual, uni-drum control for a high-speed technical exercise. Finally, in the real-world, we play songs with a drum pad and cymbal.}
    \label{fig:tasks}
    \vspace{-2mm}
\end{figure}
\section{Setup}
\begin{table*}[]
\vspace{3mm}
\caption{\textbf{Reward functions and curriculum.} \textnormal{ $n_{\text{contacts}}$ is the number of fingertip contacts, $\mathbf{p}_{\text{thumb}}, \mathbf{p}_{\text{index}}, \mathbf{p}_{\text{stick}}$ are positions, $g(\cdot)$ is a shaping function from~\cite{robopianist2023} mapping distances to $[0,1]$,  and $\mathds{1}$ denotes indicator functions for grasp. }
}
\label{tab:reward_design}
\renewcommand{\arraystretch}{1.5} 
\centering
\resizebox{\textwidth}{!}{
\begin{tabular}{lllc}
\toprule
Reward & Formula & Explanation & Weight \\
\midrule

\multicolumn{4}{c}{\textbf{\textit{In-Hand Contact}}} \\
\rowcolor[gray]{0.95} Fingertip & $\exp{(-1 / (n_{\text{contacts}} + \varepsilon))}$ & Make fingertips contact the stick & 1.0  \\

Fulcrum & $g\bigl((||\mathbf{p}_{\text{thumb}} - \mathbf{p}_{\text{stick}}||_2 + ||\mathbf{p}_{\text{index}} - \mathbf{p}_{\text{stick}}||_2) / 2 \bigr)$ & Position thumb and index finger to hold the fulcrum & 1.0\\

% \rowcolor[gray]{0.95} Hand Pose & $(\mathbf{x}_{\text{palm}} \cdot \mathbf{v}_{\text{right}} + \mathbf{z}_{\text{palm}} \cdot \mathbf{v}_{\text{down}}) / 2$ & Maintain a stable palm orientation in the world frame  & 0.5 & 0.0 \\

  Arm & $\lVert \tau_{\text{arm\_joints}} \rVert +\lVert v_{\text{arm\_joints}} \rVert $ & Penalize arm movement to incentivize finger movement & 0.03\\
\midrule 
\multicolumn{4}{c}{\textbf{\textit{External Contact}}} \\
\rowcolor[gray]{0.95} Trajectory &{$\mathds{1}_{\text{is\_grasped}} \cdot g(||\mathbf{p}_{\text{stick}} - \hat{\mathbf{p}}_{\text{stick}}||_2)$} &Guide the stick with a reference trajectory & 2.0 \\
% Grasp & $\mathds{1}_{\text{is\_grasped}}$ & Check grasp with the palm-stick distance. &   1.0  \\

\rowcolor[gray]{0.95} Contact Curriculum& N/A & Disable stick-drum contact for first N steps & N/A \\

\midrule
\multicolumn{4}{c}{\textbf{\textit{Task}}} \\
\rowcolor[gray]{0.95} Drum Hit & $\mathds{1}_{\text{is\_stick\_hit\_drum}} \cdot \mathds{1}_{\text{is\_hit\_window}}$  & Hit drum according to music &1.0 \\

\bottomrule
\end{tabular}
}
\end{table*}

\subsection{Problem Statement}
We would like to learn a robot drumming policy via reinforcement learning. Under this framework, our objective is to learn a policy $\pi(a\mid s)$ that maximizes the expected discounted cumulative reward across trajectories. We assume an environment with observation $o_t, r_t$ per timestep $t$, and discount factor $\gamma$.

In our drumming environment, we assume either a unimanual with $n_{\text{hand}}=1$, or bimanual environment with $n_{\text{hand}}=2$, where $n_{\text{hand}}$ corresponds to the number of hands. We detail the policy inputs in \cref{tab:obs} and reward functions in \cref{tab:reward_design}.

\subsection{Drum Environment}
We create a simulated drum environment in the ManiSkill framework~\cite{taomaniskill3} that consists of a bimanual robot setup and a full drum set (snare, tom, ride, hi-hat, and crash). In particular, this requires us to control and coordinate two arms and hands under a single policy, that can simultaneously play different drums. We design three types of tasks for evaluation, as shown in Fig.~\ref{fig:tasks}: (1) bimanual full–drum set songs in simulation, (2) single-drum tasks that emphasize dexterity in both simulation and the real world, and (3) bimanual two-drum songs in the real world.

\section{System Overview}
We present \fullname, a bimanual dexterous manipulation policy capable of playing drums in both simulation and the real world. An overview of the framework is shown in Figure~\ref{fig:main}. Our method adopts a hierarchical, object-centric architecture inspired by prior work~\cite{debakker2025scaffoldingdexterousmanipulationvisionlanguage, chenobject}, consisting of a high-level policy (Sec.~\ref{sec:high_level}) and a low-level dexterous policy (Sec.~\ref{sec:low_level}).
To address the complex contact dynamics inherent in drumming, we introduce contact-targeted rewards that handle both in-hand object interactions and external drumstick–surface contacts.

\subsection{High-Level Policy}
\label{sec:high_level}

The first challenge in drumming is coordinating large arm movements to navigate between drums while maintaining controllable stick motion. We formulate this as an object-centric task, where the drumstick trajectory serves as the primary reference for planning. Given the target drum sequence, we first generate the desired stick trajectory in task space. This trajectory is then transformed into corresponding end-effector motions via relative pose offsets, enabling arm-level motion planning.

However, purely kinematic planning can fail when the required motions are highly dynamic or involve rapid transitions between drums. In such cases, tracking errors accumulate and the stick may deviate from the desired trajectory. To improve robustness, we introduce a residual reinforcement learning (RL) policy that performs small corrective adjustments on top of the nominal planner. This residual formulation significantly reduces the effective action space while enabling fine-grained trajectory correction for precise stick tracking.

\subsection{Low-Level Dexterous Policy}
\label{sec:low_level}

The second challenge lies in achieving highly dexterous finger control for stable in-hand manipulation during dynamic impacts. We design the reward function with three components: (1) \textit{In-Hand Contact Rewards} for finger---stick interaction  (2) \textit{External Contact Rewards} for stick---drum interaction and (3) \textit{Task Rewards} for drum-playing. 

\textbf{In-Hand Contact Rewards.} In-hand control for the drum stick is paramount to drum playing. To enable this, we incorporate rewards targeting in-hand contact. First, we encourage finger-stick contact through a fingertip contact reward, a general reward function used in prior work~\cite{debakker2025scaffoldingdexterousmanipulationvisionlanguage} for finger-object interactions. 
Next, following human drum priors, we introduce the fulcrum reward, which specifically encourages the thumb and index finger to grasp the ``fulcrum,'' the center of the drum stick, following human priors in drumming. This further enhances finger-stick contact based on drumming priors. The fingertip contact reward encourages the fingers to successfully touch and grasp the object, which in this case, allows us to hold and manipulate the stick. Next, we propose an arm penalty constraint, which reduces excessive arm movements, making in-hand contact more natural. This prevents the agent from manipulating the stick with arm movements, incentivizing the agent to develop fine-grained finger control. 
Energy minimization has been widely used in locomotion~\cite{tan2018simtoreallearningagilelocomotion,9341571,fu2021minimizing} to induce diverse gaits, and here, we adapt it for in-hand contact with the stick.

\textbf{External Contact Rewards.} 
To address external contact, we divide the challenges into two stages: (1) initiating contact
% (2) managing exploration under contact, 
and (2) maintaining long-horizon contact. We address the first with a trajectory reward and contact curriculum, and we address the second with a reactive grasp reward. 

First, we would like our policy to successfully initiate external contact between the stick and drum. To achieve this, we apply a trajectory reward, which explicitly guides the speed and motion, hence guiding precise control of the contact force. In our case, we pre-compute the desired trajectories of the drumstick tip and end by modeling drum hits with a sinusoidal wave and interpolating between drum positions across hits. This approach is inspired by prior work on high-level trajectory planning and low-level control~\cite{debakker2025scaffoldingdexterousmanipulationvisionlanguage,qian2024pianomime,chenobject,lum2025crossinghumanrobotembodimentgap}. In this drumming environment, both arm actions and hand actions can help optimize this objective. For instance, when transitioning in between drums, both hand and arm control are important for following the stick trajectory. However, closer to the drum, this primarily becomes a dexterous challenge, as the fingers should be initiating the contact with the drum, compared to excessive, large arm movements.

While this reward guides the hitting motion of the stick, the reward by itself is often insufficient to learn how to properly hit the drum.
Specifically, during learning, the stick often rests on the drumhead, which often blocks the exploratory finger motions necessary for controlling the stick.
To address this issue, we introduce a contact curriculum. Initially, contact between the stick and the drum is disabled, allowing the agent to practice trajectory following in free space while following the trajectory reward.  
Contact is later reintroduced, enabling the policy to more effectively initiate drum hits. 
This curriculum helps the policy learn to initiate external contact, as we decompose the problem into first following a motion, and then learning reactive behaviors to continue following the trajectory with the external forces.
This curriculum shares similarities with DexMachina~\cite{mandi2025dexmachinafunctionalretargetingbimanual}, which uses virtual object controllers to prevent early failures caused by gravity. In contrast, our curriculum targets contact-related randomness -- a more challenging source of instability -- and is simpler, requiring no modifications to object assets.

\textbf{Task Rewards.} Similar to prior works~\cite{shahid2025drummer}, we add a sparse hit reward to check whether the drum is hit at a specified time. 

\section{EXPERIMENTS}
We seek to answer the following questions:
\begin{enumerate}
    \item Is dexterity \emph{essential} for achieving robust long-horizon drumming? Can we learn such a dexterous policy? 
    \item How can we enable finger-driven control for precise drumming? 
    \item How do our design decisions affect dexterous control? 
    \item Can the learned behaviors transfer to real world?
\end{enumerate}

\begin{table}[]
\vspace{3mm}
\caption{\textbf{Observation space.}\textnormal{ $L$ denotes lookahead horizon}.}
\label{tab:obs}
\centering
\resizebox{0.6\linewidth}{!}{
\begin{tabular}{ll}
\toprule
Observation & Dimension \\
\midrule
Arm Proprioception & $ 7 \times n_{\text{hand}} $ \\
Hand Proprioception & $ 20 \times n_{\text{hand}}$ \\ 
\midrule
Stick Head Proprioception & $ 3 \times n_{\text{hand}}$ \\ 
Stick Tail Proprioception & $ 3 \times n_{\text{hand}}$ \\ 
Trajectory Plan: Stick Head & $ 3 \times n_{\text{hand}} \times L$ \\ 
Trajectory Plan: Stick Tail & $ 3 \times n_{\text{hand}} \times L$\\ 
\midrule
Stick is Grasped & $ 1 \times n_{\text{hand}}$ \\ 
Previously Played Drum & $ 7 $ (discrete) \\ 
Next Drum to Play &$7$ (discrete) \\ 
Time Before Next Drum Hit & $1$ \\
\bottomrule
\end{tabular}
}
\end{table}
\vspace{-1mm}

\subsection{Experimental Setup}
\subsubsection{Hardware Setup} We use two 7-DOF Franka Panda arms and two 20-DOF Tesollo DG-5F hand in both simulated bimanual and real-world unimanual tasks. We run policy inference at 20 Hz, and we use a PID joint position controller that runs at 100 Hz. 

We showcase real-world drumming with a unimanual, drum pad and cymbal setup. We create a digital twin of the real world by matching the drum, cymbal, and stick positions and sizes. We apply randomization to improve robustness for sim-to-real. 

To track the stick proprioception, we paint the end of the drumstick. We use color segmentation, depth readings from a RealSense camera, camera intrinsics, and camera extrinsics to project it into robot frame.

\begin{figure}
    \centering
    \vspace{2mm}\includegraphics[width=0.85\linewidth]{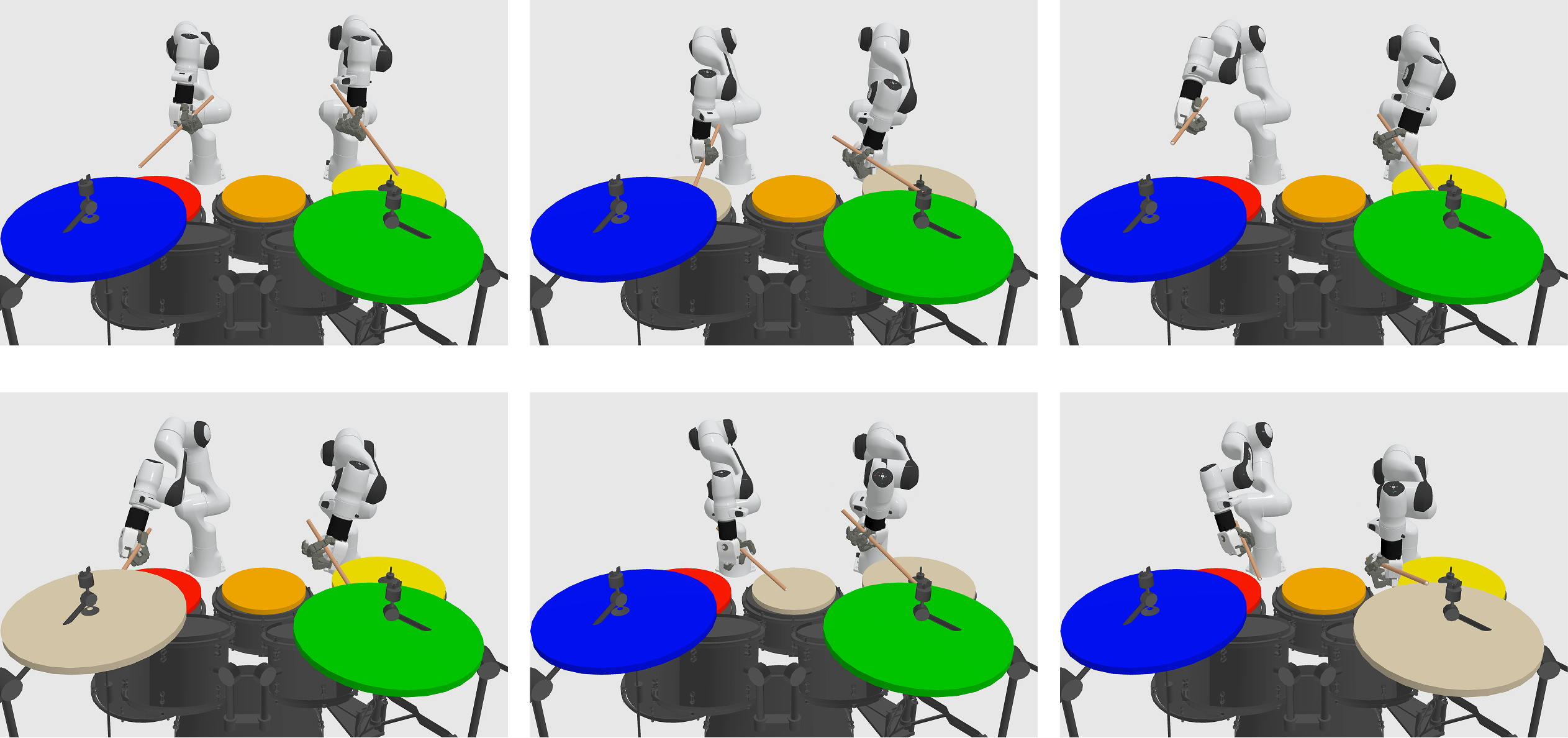}
    \caption{\textbf{Bimanual Song-Playing Rollout.} We visualize 6 frames across a single song trajectory, with lighter colored drums and cymbals corresponding to a hit. Every song requires multiple combinations of drums to be hit.}
    \vspace{-2mm}
    \label{fig:bimanual_rollout}
\end{figure}

\begin{figure*}
\vspace{6mm}
    \centering
    \includegraphics[width=0.85\linewidth]{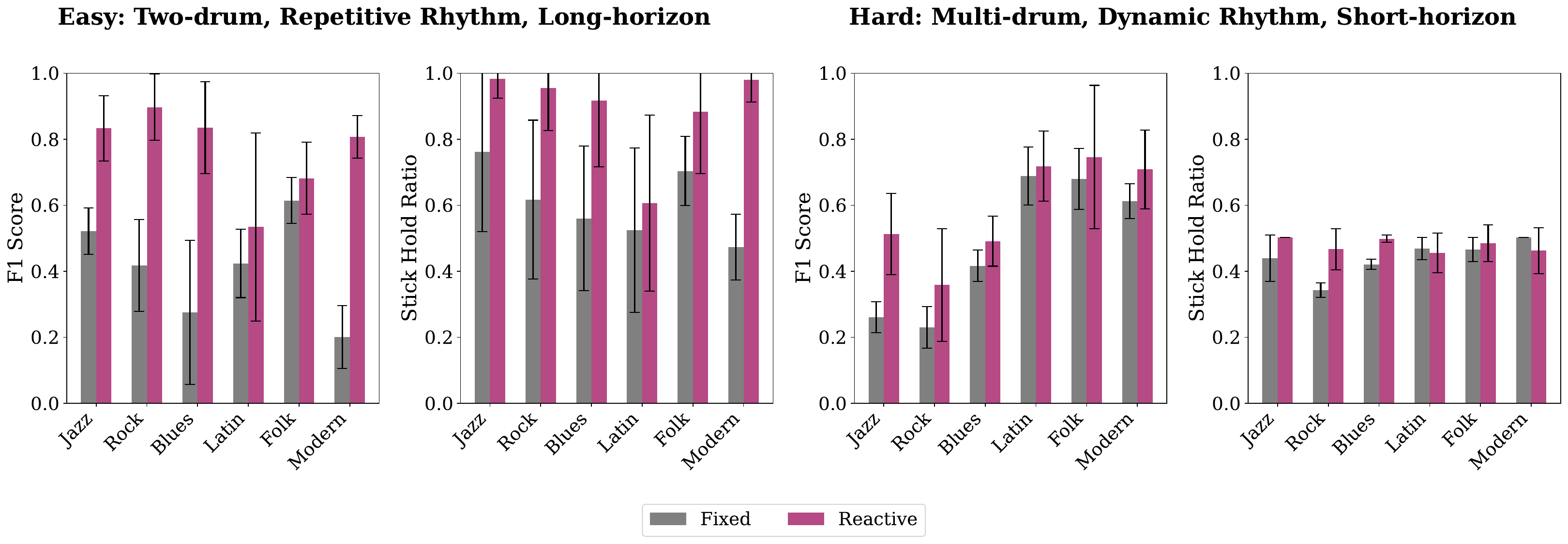}
    \vspace{-10pt}
    \caption{\textbf{Results for Dexterous Song-Playing.} \textbf{\emph{Left}}: Reactive grasp outperforms fixed grasp by a large margin in long-horizon contacts. \textbf{\emph{Right}}: For more challenging songs requiring frequent drum-to-drum transitions, reactive grasp still improves performance, but with a smaller margin, primarily due to the reduced action space of fixed grasp.}
    \label{fig:bimanual_all}
\end{figure*}

\begin{figure*}
% \vspace{8mm}
    \centering
    \includegraphics[width=0.85\linewidth]{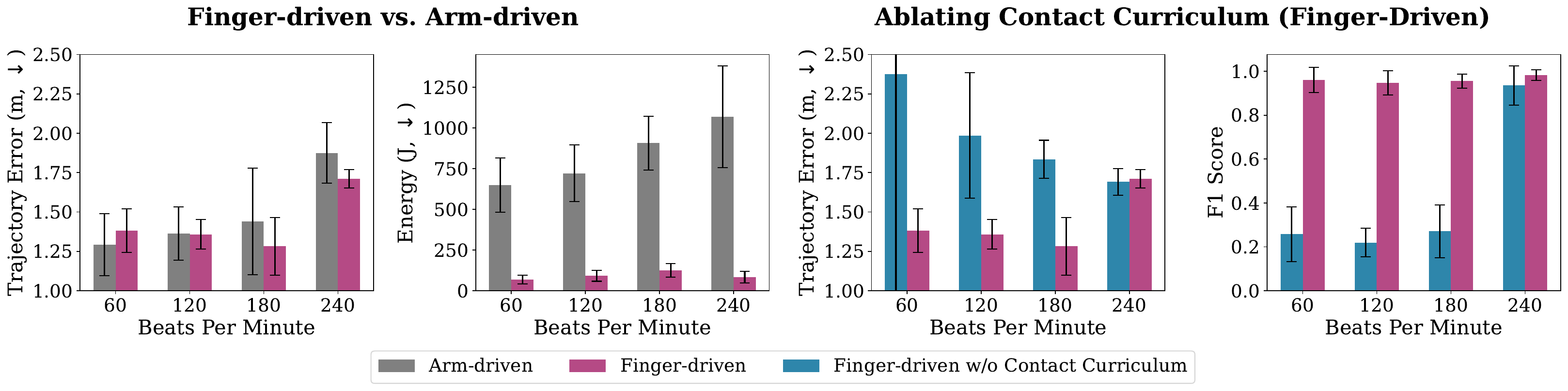}
    \vspace{-10pt}
    \caption{\textbf{Results for Finger-Driven Control.} \textbf{\emph{Left:}} As tempo (beats per minute) increases, trajectory error for finger-driven control decreases and the gap to arm-driven control widens, showing the superior dexterity of finger-driven control. Energy consumption is also substantially lower compared to arm-driven control. \textbf{\emph{Right:}} Without contact-targeted rewards, finger-driven control struggles to manage contact interactions effectively.}
    \label{fig:dex_all}
\end{figure*}

\subsubsection{Evaluation Metrics}

For bimanual song playing, we evaluate performance using two metrics: (1) the F1 score for song performance, and (2) the stick-hold ratio, defined as the fraction of time the stick remains held in hand across the total duration, which reflects the effectiveness of in-hand control. For dexterous fine control, we use (1) trajectory error, measuring how accurately the policy follows fine-grained trajectories, and (2) energy consumption, capturing the overall efficiency of the system.

\subsubsection{Drum Songs} We use MIDI files to specify drum sequences, following prior work~\cite{robopianist2023, shahid2025drummer}. 
% The songs are selected from an open-access MIDI drum dataset and cover multiple musical genres. To ensure feasibility for robotic control, we adjust the tempo and extract short playable segments.
For simulation experiments, we construct \textbf{Easy} and \textbf{Hard} tracks from an open-source MIDI drum dataset covering six different genres, using a bimanual setup with multiple drums. Easy tracks consist of repetitive rhythmic patterns designed to evaluate long-horizon stability, while Hard tracks involve more complex rhythms requiring coordinated hits across different drums.

We also design a Finger-Driven Control task that focuses on high-speed single-drum playing to test fine-grained finger dexterity.

% For real-world evaluation, we use two songs—\textit{Seven Nation Army} and \textit{Everlong}—performed on a simplified setup with two drums (snare and hi-hat). The slower song emphasizes larger arm movements, while the faster one requires higher finger dexterity. Additional details on song selection and processing are provided in the Appendix~\ref{app:song_process}.

\subsection{Dexterity for Bimanual Song Playing}
\label{exp:bimanual_song}
In our first set of experiments, we evaluate the importance of dexterity for long-horizon tool manipulation. For drumming, grasping the drum stick and repeatedly hitting a drum for extended periods of time likely necessitates reorientation and adjustment of the drum stick. To test the effectiveness of our dexterous policy, we compare \textit{Fixed Grasp}, where the finger joints are frozen after reaching an initial grasp of the stick, to our method, \textit{Reactive Grasp}, where the agent exerts dexterous control of the stick.

We evaluate the F1 Score and Hold Duration of the stick. The F1 Score represents how well the policy can play the song, showing that \fullname{} can learn a successful dexterous drumming policy. The Hold Duration evaluates how \textit{Fixed Grasp} and \textit{Reactive Grasp} are affected by slippage as the robot continues to move and hit the drum stick.

In ~\cref{fig:bimanual_all}, we present two setups: an easier scenario with a repetitive loop but long horizon (\emph{left}), and a more challenging scenario with multiple drums, dynamic rhythm, but shorter horizon (\emph{right}). In the left case, the reactive grasp clearly outperforms the fixed grasp in both song performance and stick hold duration, highlighting the necessity of reactive, closed-loop dexterous control for long-horizon contact. In the right case, the reactive grasp still outperforms the fixed grasp, but with a much smaller margin. This is mainly because the fixed grasp only needs to learn arm motion within a lower-dimensional action space, which makes it easier to handle complex drum-to-drum transitions. The result shows a trade-off between dexterity and learning complexity, making it an interesting direction for balancing the two.

\subsection{Finger-Driven Control}
\label{exp:finger_control}
Next, we explore how to enable fine-grained, dexterous motions instead of relying on unnatural, whole-arm movements. Drumming can be attempted through arm movements or dexterous finger movements, but the precise finger control can often be overshadowed by initial arm exploration and movement. In this experiment, we evaluate how \fullname{} guides dexterity, and whether this leads to better performance of the song. 

To evaluate dexterity, we choose an exercise that requires fine-grained finger movements. This exercise requires playing a single drum very rapidly, which is difficult for arm movements to follow and motivates dexterous finger control. To enable finger-driven control, we incorporate the arm energy penalty and contact curriculum, with reward terms listed in~\cref{tab:reward_design}. The arm energy penalty limits arm movement in favor of finger movements, and the contact curriculum helps the finger-driven policy learn to handle external contact. The arm-driven policy optimizes the same reward terms, with the exception of the arm penalty and contact curriculum.

For this task, we evaluate the F1 Score, Trajectory Error, and Energy Consumption across a range of speeds for the exercise, denoted by Beats Per Minute. The F1 Score represents how well the policy plays the song. The Trajectory Error shows, more precisely, how well the drum stick can follow the desired trajectory. In particular, because this song requires greater rotation of the stick up and down, this captures how well arm or finger driven movements are able to reproduce this motion. Finally, high Energy Consumption implies extra, unnecessary movements, which is less desirable due to safety and sustainability concerns. We evaluate these metrics across a range of Beats Per Minute (BPM) for our exercise. A higher BPM implies quicker hits and less time in between hits, and is thus more challenging.

\begin{figure}
    \centering
    \vspace{2mm}
    \includegraphics[width=0.85\linewidth]{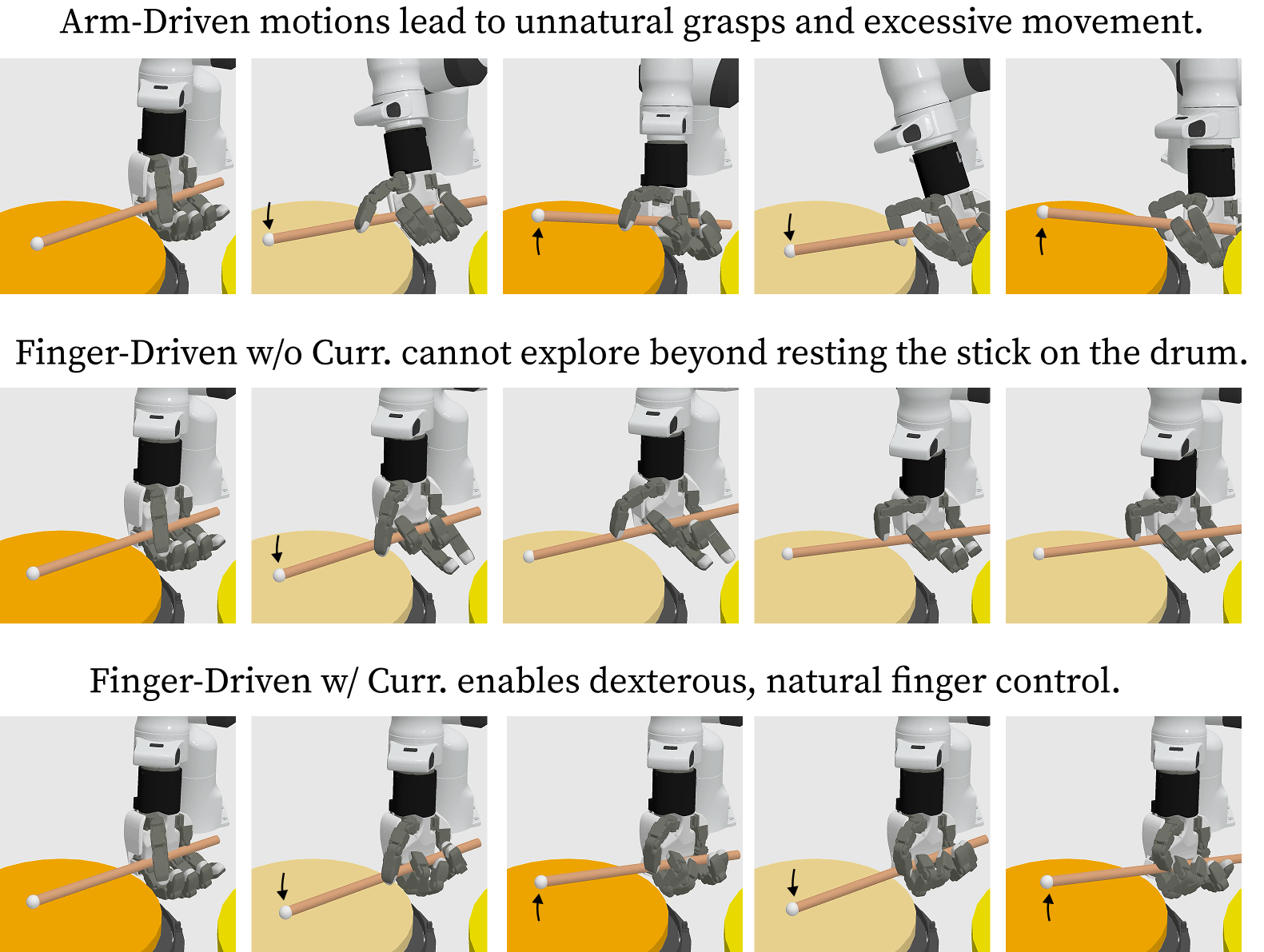}
    \caption{\textbf{Sample Rollouts for the Dexterous Exercise.} For fast-playing exercises, the arm-driven policy moves to unnatural positions, leading to energy-intensive and potentially dangerous positions. Finger-driven control without the contact curriculum is unable to effectively learn dexterous control, whereas adding the contact curriculum leads to the most natural, effective drum hits. The lighter-colored drum head denotes that the drum is being hit by the stick.}
    \label{fig:dext_rollouts}
    \vspace{-10pt}
\end{figure}

As shown in ~\cref{fig:dex_all}, finger-driven control outperforms arm-driven control, particularly as the BPM increases. This highlights the necessity of fingers for fine-grained, high-speed motions, as arm-driven motion is unable to accurately replicate the quick stick movements. Moreover, finger-driven motion results in significantly lower energy consumption, which is advantageous for practical deployment. Qualitatively, in \cref{fig:dext_rollouts}, we find finger-driven motions to look more natural and human-like, whereas arm-driven motions are clunky and dangerous.

The ~\cref{fig:dex_all} (right) investigates what enables this dexterous finger movement. We ablate our finger-driven motion policy by removing the contact curriculum. Across different BPMs, removing the contact curriculum leads to much higher trajectory errors, implying that the curriculum is necessary to learn finger dexterity for following the stick trajectory. An exception occurs at a very high tempo (240 BPM), where the robot only needs to lift the stick slightly above the drum before hitting it again. Here, the effect of the contact curriculum is limited because the dexterous motion is small and easier to learn. This is further exemplified through the F1 scores, which shows a drastic gap when the contact curriculum is not enabled, besides for the fastest tempo (240 BPM). Qualitatively, in \cref{fig:dex_all}, without the contact curriculum, the stick often rests on the drum head, preventing effective finger exploration. We hypothesize that because finger movements are relatively small, they can easily be canceled out by external contact. 

\begin{figure}
\vspace{4mm}
    \centering
\includegraphics[width=0.83\linewidth]{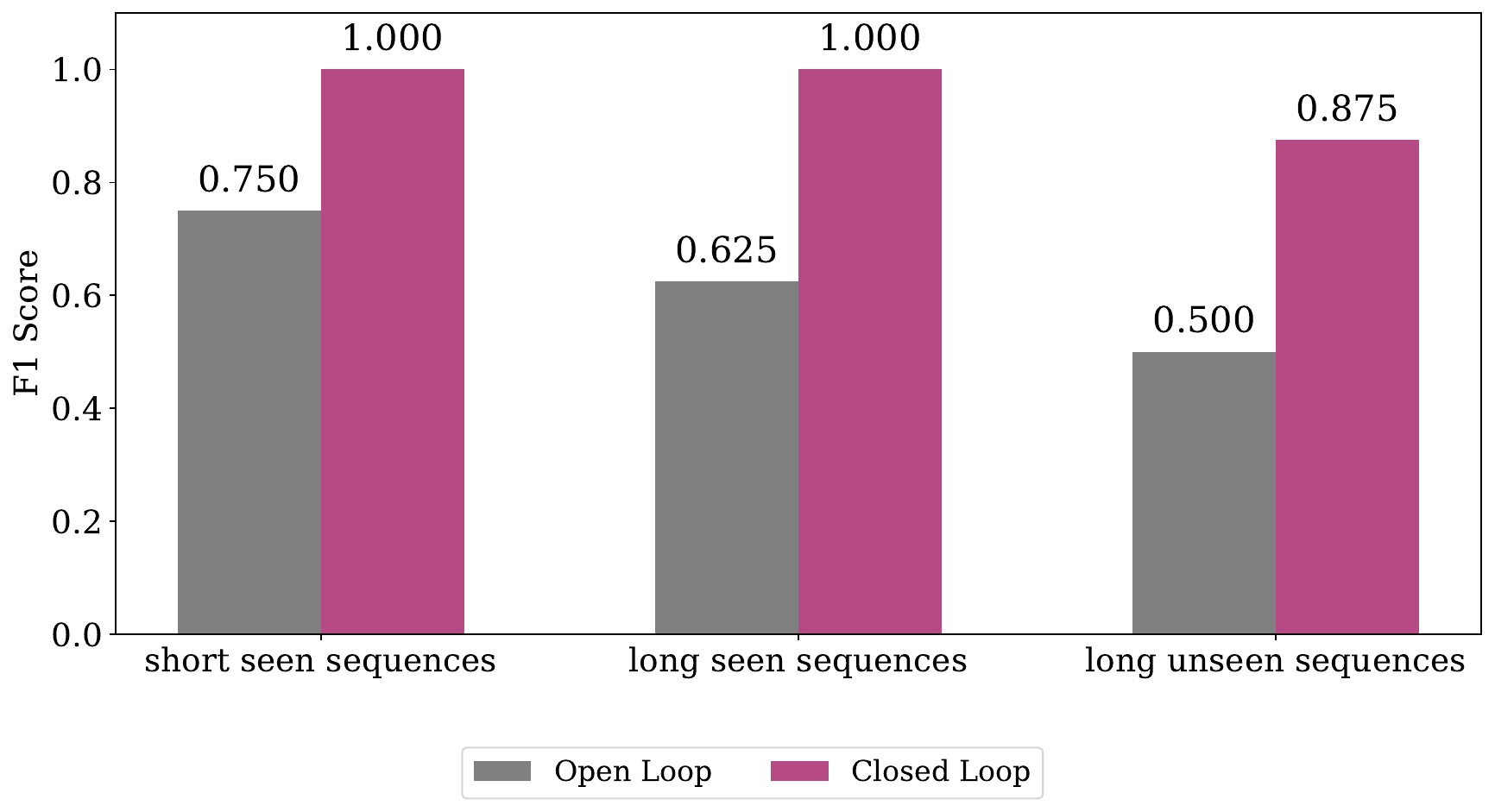}
    \vspace{-10pt}
    \caption{\textbf{Results for Real-world Unimanual Drum Playing.} We denote the cymbal as $c$ and the drum pad as $d$. The short seen sequence used for training is $ccdd$. The long seen sequence is $ccccdddd$, while the long unseen sequence is $ccddccdd$. Open-loop replay achieves reasonable performance but cannot adapt to real-world dynamics. Our policy is able to play unseen sequences (e.g., $d \rightarrow c$) by conditioning on the desired trajectory.}
    \label{fig:cymbal_snare}
    \vspace{-5mm}
\end{figure}
\subsection{Ablations for Arm Control}

We study the contribution of each component in the arm control pipeline through an ablation analysis of motion planning and residual RL. As shown in the accompanying videos, the full system that combines motion planning with residual RL achieves precise and stable drumming performance, reaching an F1 score of 1.0. When removing the residual RL policy, the system relies solely on kinematic planning and fails to perform fine-grained corrections during drum contact. This results in inaccurate hits, limited adaptability in finger control, and a reduced F1 score of 0.8. Further removing motion planning and training a policy purely with RL from scratch significantly degrades performance. In this setting, the policy struggles to coordinate large-scale arm movements with dexterous hand control, leading to frequent missed hits and an F1 score of 0.5. These results highlight that motion planning provides a strong structural prior for global coordination, while residual RL is critical for local correction and precise stick control.

\subsection{Real-World Drumming}
\textbf{Open-Loop vs. Closed-Loop Drumming for Robustness (Unimanual, Two-Drum). }Finally, we show results on real-world drumming with a drum pad and cymbal. We evaluate the effectiveness of our sim-to-real policy across multiple songs in an open and closed-loop setting, similar to~\cite{zeulner2025learning}. First, we evaluate F1 performance on the seen song our RL agent is trained on. Then, we evaluate on an extended version of the song, which should have hits and trajectories that are seen in the train song. Lastly, our most difficult song not only requires more hits than the train song, but it also includes unseen drum transitions. This is to evaluate the generalization of our policy to out-of-distribution songs. For our open-loop setting, we directly run the policy in simulation and replay actions, whereas for closed-loop song-playing, we run policy inference based on real-world states. 

In \cref{fig:cymbal_snare}, we find that our closed-loop policy consistently outperforms the open-loop policy. This shows that closed-loop control is essential for real-world drumming, as the robot is able to react to stick movements to better play the song. Our closed-loop policy is able to play both the train song and an extended version of the train song with an F1 score of 1.0, showing that sim-to-real can enable effective drum playing. Additionally, the closed-loop policy can play songs with unseen drum transitions (e.g., from drum pad to cymbal), which may be due to high-level trajectory guidance that our policy is conditioned on~\cref{tab:obs}. 

\begin{figure}
    \centering
    \vspace{2mm}    \includegraphics[width=0.85\linewidth]{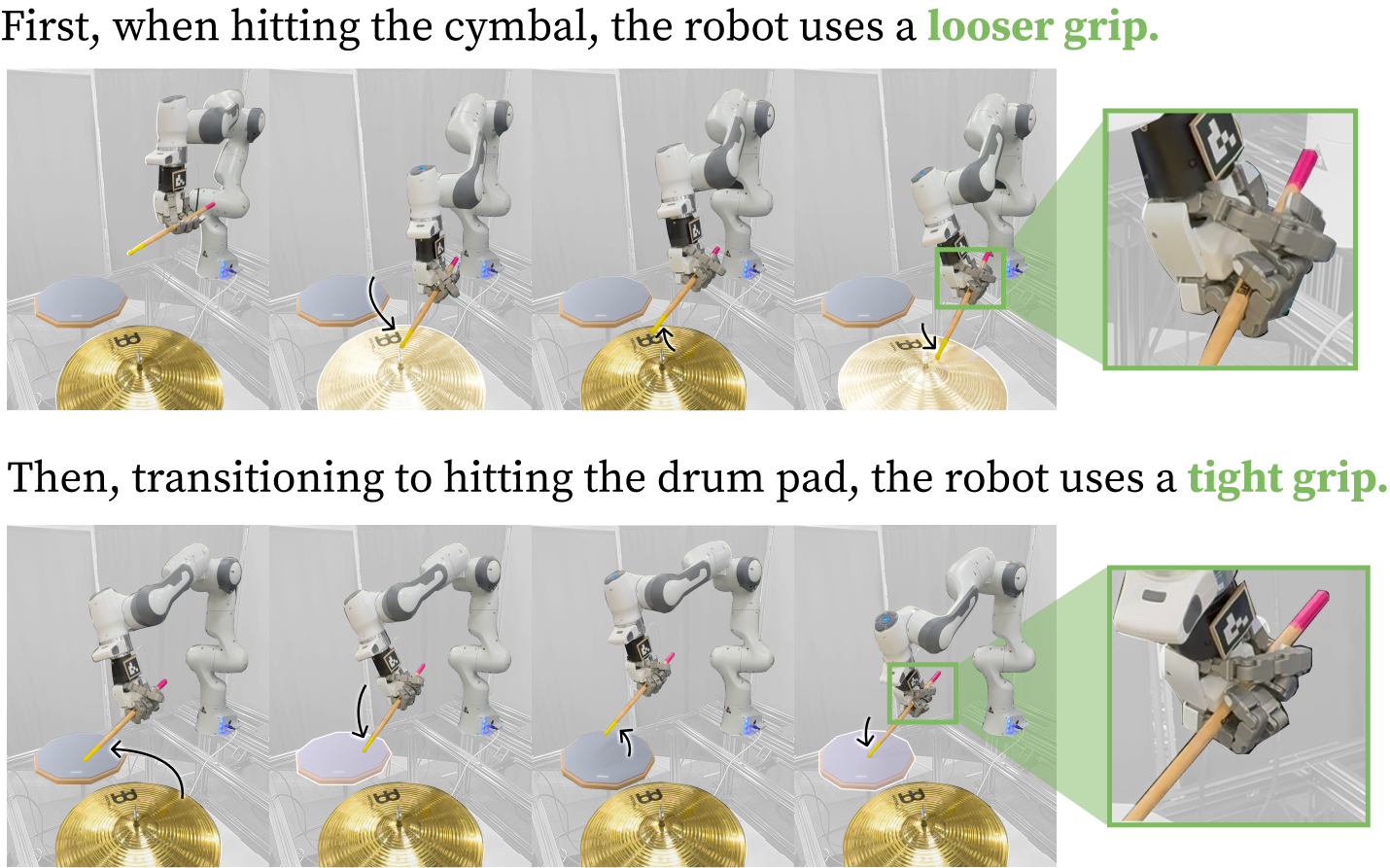}
    \caption{\textbf{Real World Rollouts.} We visualize part of one closed-loop trajectory. The lighter colors for the cymbal and drum show when the robot hits them with the stick, and the arrows visualize the direction the stick was moving in. In this continuous rollout, we two cymbal hits and two drum hits. Notably, for the cymbal, which is not fixed and can rotate around the world vector, the robot strikes with a relatively loose grip (top). However, after hitting the drum pad, which is stable, the hand adjusts and forms a firmer grip (bottom), as shown by the movement of the index finger and thumb.}
    \vspace{-5mm}
    \label{fig:real_rollout}
\end{figure}

In Fig.~\ref{fig:real_rollout}, we show a sample rollout from real-world drum playing. Notably, we highlight that the dynamics of hitting a cymbal and a drum pad differ: the cymbal is not fixed and involves less contact, whereas the drum pad is fixed and introduces more substantial contact. When striking the cymbal, the grip remains relatively loose since the interaction does not cause significant instability. In contrast, after striking the drum pad, the grip becomes firmer to handle the stronger contact. This adaptation shows the effectiveness of the reactive grasp.

\textbf{Dexterous Control on Exercise Songs (Unimanual, Single-Drum).}  
We also showcase the experiments from~\ref{exp:finger_control} in the real world. As shown in the supplemental video, DexDrummer is able to execute highly dexterous actions to play at high speeds. Qualitatively, we evaluate the behavior under different speeds. At 120 beats per minute (BPM), the last three fingers are used to drag the stick downward for the upstroke. At 240 BPM, the last three fingers instead press the stick to generate a faster downstroke. This behavior resembles the technique used by human drummers.

% \textbf{Real Song Playing (Bimanual, Two-Drum).}  Finally, we demonstrate playing two real modern songs: \textit{Everlong} by Foo Fighters (with the speed slowed down by a factor of two) and \textit{Seven Nation Army} by The White Stripes. For each song, we extract a segment of approximately 15 seconds. The results shown in the videos demonstrate that our framework is able to play real, bimanual songs robustly.

\section{CONCLUSION}
We introduced \fullname{}, a testbed for dexterous manipulation that unifies the challenges of in-hand control, contact-rich interaction, and long-horizon tasks. 
To address these coupled challenges, we propose a hierarchical framework in which a high-level policy leverages parameterized motion primitives to reduce RL exploration, while a low-level policy handles both in-hand and external contacts under dexterous control.
We demonstrated the effectiveness of our framework through bimanual song playing and fine-grained control on the real robot.
In future work, we aim to extend these insights to broader problems that involve interaction with in-hand objects and the physical world.

\textbf{Limitations and Future Work.}
We are excited about future directions for dexterous drumming. For one, \fullname cannot play multi-drum songs at human speed, and we slow down our songs in order to make it feasible. Future directions may explore how to play to drum tracks in real-time. Similarly, current experiments test performance for up to 400 timesteps (20 seconds), whereas real-world songs are often 3-5 minutes. Improving speed, robustness, and control is paramount to improved song performances.

For our real-world experiments, we currently only show a bimanual performance with two drums. Future work may showcase dexterous drumming with a full drum set. Furthermore, we do not incorporate sim-to-real techniques besides domain randomization, and further research into reducing the sim-to-real gap may lead to stronger song-playing.

\section{Acknowledgment}

This work was funded by ONR MURI N00014-25-1-2479, and ONR YIP N00014-22-1-2293, NSF \# 2218760 and NSF \#1941722. We would like to thank the ILIAD members for their feedback on an early version of the paper. We thank Satvik Sharma and Vincent de Bakker for hardware and software advice, Suvir Mirchandani for hardware assistance, and Jayson Meribe and Kushal Kedia for helpful discussions.

%%%%%%%%%%%%%%%%%%%%%%%%%%%%%%%%%%%%%%%%%%%%%%%%%%%%%%%%%%%%%%%%%%%%%%%%%%%%%%%%

% \addtolength{\textheight}{-12cm}   % This command serves to balance the column lengths
                                  % on the last page of the document manually. It shortens
                                  % the textheight of the last page by a suitable amount.
                                  % This command does not take effect until the next page
                                  % so it should come on the page before the last. Make
                                  % sure that you do not shorten the textheight too much.

%%%%%%%%%%%%%%%%%%%%%%%%%%%%%%%%%%%%%%%%%%%%%%%%%%%%%%%%%%%%%%%%%%%%%%%%%%%%%%%%

%%%%%%%%%%%%%%%%%%%%%%%%%%%%%%%%%%%%%%%%%%%%%%%%%%%%%%%%%%%%%%%%%%%%%%%%%%%%%%%%

%%%%%%%%%%%%%%%%%%%%%%%%%%%%%%%%%%%%%%%%%%%%%%%%%%%%%%%%%%%%%%%%%%%%%%%%%%%%%%%%
% \section*{APPENDIX}

% Appendixes should appear before the acknowledgment.

% \section*{ACKNOWLEDGMENT}

% The preferred spelling of the word ÒacknowledgmentÓ in America is without an ÒeÓ after the ÒgÓ. Avoid the stilted expression, ÒOne of us (R. B. G.) thanks . . .Ó  Instead, try ÒR. B. G. thanksÓ. Put sponsor acknowledgments in the unnumbered footnote on the first page.

%%%%%%%%%%%%%%%%%%%%%%%%%%%%%%%%%%%%%%%%%%%%%%%%%%%%%%%%%%%%%%%%%%%%%%%%%%%%%%%%

% References are important to the reader; therefore, each citation must be complete and correct. If at all possible, references should be commonly available publications.

\bibliographystyle{IEEEtran}
\bibliography{references}

\newpage
\section*{Appendix}

The code is available at \url{https://github.com/hc-fang/dexdrummer} , and the videos are available at \url{https://dexdrummer.github.io/} .

\subsection{Real-World Bimanual Song-Playing}
We further showcase a real-world bimanual drum performance on our website. We use two songs—\textit{Seven Nation Army} and \textit{Everlong}—performed on a simplified setup with two drums (snare and hi-hat). The slower song emphasizes larger arm movements, while the faster one requires higher finger dexterity. The results shown in the videos demonstrate that our framework is able to play real, bimanual songs robustly.

\subsection{Song Processing}

\label{app:song_process}

Like prior works~\cite{robopianist2023, shahid2025drummer}, we specify songs by importing MIDI files into the environment. MIDI is a widely used representation that encodes instruments and timing in a concise, discrete format. 

\paragraph{Bimanual Song Playing.} We obtain MIDI songs from an open-access website\footnote{\href{https://mididrumfiles.com/tag/midi-files/}{https://mididrumfiles.com/tag/midi-files/}}
. From this collection, we select six different genres of music, and for each track, we extract both an Easy and Hard segment. To ensure playability for the robot policy, all songs are slowed down by a factor of three. 
The Easy tracks feature repetitive loops, which we evaluate over 400 timesteps (20 sec) to test long-horizon control. The Hard tracks require hitting multiple drums with dynamic rhythms, and we evaluate these with 200 timesteps (10 sec). All tracks are performed using a bimanual setup.

\paragraph{Finger-Driven Control Experiment} For our Finger-Driven Control experiment (Section~\ref{exp:finger_control}), we use an exercise that emphasizes finger dexterity by playing a single drum at a very high speed of up to 240 beats per minute (i.e., 4 hits per second). We train with 100 timesteps in a uni-manual setup.

\paragraph{Real-world Bimanual  Song Playing} For real-world bimanual playing, we select two songs: \textit{Seven Nation Army} by The White Stripes, with approximately 120 beats per minute (BPM), and \textit{Everlong} by Foo Fighters, a faster and more dexterous piece with approximately 160 BPM. The first song primarily emphasizes larger arm movements, whereas the second requires greater finger dexterity due to its higher tempo and denser rhythmic patterns.

To simplify the physical setup, we restrict the drum configuration to only a snare and a hi-hat and adapt the drum sheet accordingly. Specifically, notes corresponding to left or right cymbals are mapped to the hi-hat, while notes corresponding to toms are mapped to the snare.

\subsection{Simulation Training}

We train our policies with Proximal Policy Optimization~\cite{schulman2017proximalpolicyoptimizationalgorithms}, running for 60M steps on bimanual tasks and 40M steps on unimanual tasks. The training setup uses a discount factor of $\gamma = 0.8$,  and 1024 parallel environments. We adopt generalized advantage estimation (GAE) with $\lambda = 0.9$ and employ a 3-layer MLP policy network with hidden dimensions of size 512. For the contact curriculum, contact between the drum stick and the drum pad is disabled during the first 10,000 steps. 

For sim-to-real training, we incorporate domain randomization. At each step, uncorrelated Gaussian noise sampled from $\mathcal{N}(0, 0.05^2)$ is added independently to the proprioception and stick positions in the observation space. The stick’s friction coefficient is perturbed with noise drawn from a uniform distribution $\mathcal{U}(-0.2, 0.2)$. In addition, control gains are scaled by a random factor sampled from $\mathcal{U}(0.9, 1.1)$ at environment initialization, and this factor is kept fixed for each environment throughout training.
% \begin{thebibliography}{99}

% \bibitem{c1} G. O. Young, ÒSynthetic structure of industrial plastics (Book style with paper title and editor),Ó 	in Plastics, 2nd ed. vol. 3, J. Peters, Ed.  New York: McGraw-Hill, 1964, pp. 15Ð64.
% \bibitem{c2} W.-K. Chen, Linear Networks and Systems (Book style).	Belmont, CA: Wadsworth, 1993, pp. 123Ð135.

% \end{thebibliography}

\end{document}